\documentclass[conference]{IEEEtran}
\usepackage{pifont}
\usepackage{graphicx}%
\usepackage{multirow}%
\usepackage{tabularx}
\usepackage{amsmath,amssymb,amsfonts}%
\usepackage{amsthm}%
\usepackage{mathrsfs}%
\usepackage{xcolor}%
\usepackage{afterpage}
\usepackage{textcomp}%
\usepackage{placeins}
\usepackage{float}
\usepackage{manyfoot}%
\usepackage{makecell}
\usepackage{booktabs}%
\usepackage{listings}%
\usepackage{amsmath,amssymb,amsfonts}
\usepackage{graphicx}
\usepackage{textcomp}
\usepackage{xurl}
\usepackage{enumitem}
\usepackage{amsthm,amssymb,amsfonts}
\usepackage{tikz,lipsum}
\usepackage[most]{tcolorbox}
\usepackage[svgnames]{xcolor}
\usepackage{xcolor}
\usepackage{hyperref}

\theoremstyle{thmstyleone}%
\theoremstyle{thmstyletwo}%
\usepackage{caption}
\theoremstyle{thmstylethree}%
\newcommand{\tinyscript}{\fontsize{6pt}{7pt}\selectfont}
\begin{document}

\title{Comparative Analysis of Large Language Models in Healthcare}

\author{
Subin Santhosh$^{\ast}$, Farwa Abbas$^{\ast}$$^{\dagger}$, Hussain Ahmad$^{\ast}$, Claudia Szabo$^{\ast}$%
\\ $^{\ast}$Adelaide University, Australia\\
$^{\dagger}$South Australian Health and Medical Research Institute, Australia
\\[0.3em]
Email:
subin.santhosh@student.adelaide.edu.au; \{farwa.abbas, hussain.ahmad, claudia.szabo\}@adelaide.edu.au
}

\maketitle

\begin{abstract}
\textit{Background:} Large Language Models (LLMs) are transforming artificial intelligence applications in healthcare due to their ability to understand, generate, and summarize complex medical text. They offer valuable support to clinicians, researchers, and patients, yet their deployment in high-stakes clinical environments raises critical concerns regarding accuracy, reliability, and patient safety. Despite substantial attention in recent years, standardized benchmarking of LLMs for medical applications has been limited.
\textit{Objective:} This study addresses the need for standardized comparative evaluation of LLMs in medical settings.
\textit{Method:} We evaluate multiple models, including ChatGPT, LLaMA, Grok, Gemini, and ChatDoctor, on core medical tasks such as patient note summarization and medical question answering, using the open-access datasets MedMCQA, PubMedQA, and Asclepius, and assess performance through a combination of linguistic and task-specific metrics.
\textit{Results:} The results indicate that domain-specific models, such as ChatDoctor, excel in contextual reliability, producing medically accurate and semantically aligned text, whereas general-purpose models like Grok and LLaMA perform better in structured question-answering tasks, demonstrating higher quantitative accuracy. This highlights the complementary strengths of domain-specific and general-purpose LLMs depending on the medical task.
\textit{Conclusion:} Our findings suggest that LLMs can meaningfully support medical professionals and enhance clinical decision-making; however, their safe and effective deployment requires adherence to ethical standards, contextual accuracy, and human oversight in relevant cases. These results underscore the importance of task-specific evaluation and cautious integration of LLMs into healthcare workflows.
\end{abstract}
% \maketitle

\section{Introduction}

\begin{figure}[t]
    \centering
    \includegraphics[trim={0 0 0 0.2cm},clip,width=0.8\linewidth]{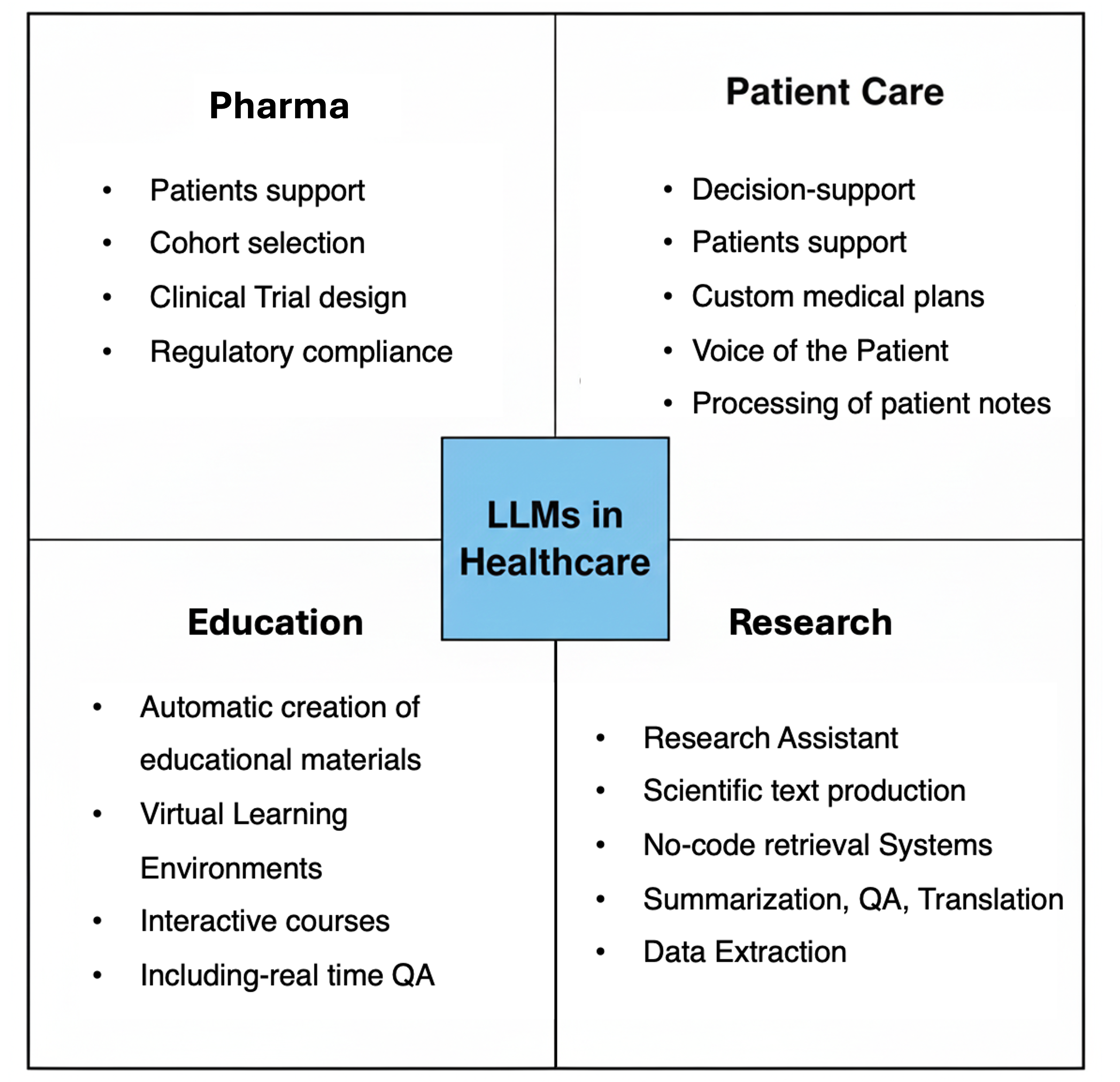}
    \centering
    \caption{Applications of LLMs in Healthcare}
    \label{LLMs_in_Healthcare}
    \vspace{-1\baselineskip}
\end{figure}

Artificial Intelligence (AI) has emerged as one of the most transformative technologies of the modern era \cite{abbas2025scalar, jois2026australian, abbas2024robust}, fundamentally reshaping a wide range of sectors, including finance \cite{zhang2025regimefolio}, education \cite{ahmad2025future}, offensive cyber operations \cite{goel2025co, goel2024machine}, defensive cyber operations \cite{jayalath2024microservice, ahmad2025survey}, and cloud computing \cite{ahmad2025towards, ahmad2024smart, ahmad2025resilient}. This transformation is driven by AI ability to automate complex decision-making processes, extract insights from large-scale data, and adapt dynamically to evolving environments \cite{ahmad2023review, abdulsatar2025towards}. In cybersecurity, for instance, AI is increasingly leveraged to both enhance sophisticated attack strategies and strengthen intelligent defense mechanisms \cite{ullah2026skills}, highlighting its dual-use nature in adversarial settings \cite{zhang2026explainable}. Among the most revolutionary advancements within AI are Large Language Models (LLMs) \cite{haque2022think}, which are capable of understanding, generating, and reasoning with human language at an unprecedented scale \cite{chopra2026chatnvd}. These models have significantly advanced natural language processing by enabling context-aware interactions, knowledge synthesis, and task generalization across domains \cite{chen20253s}. As a result, LLMs are being rapidly adopted in critical sectors such as healthcare, where they support applications ranging from clinical decision assistance and medical documentation to patient interaction and knowledge retrieval. For example, Figure \ref{LLMs_in_Healthcare} illustrates the diverse applications of LLMs across key healthcare domains.

Healthcare, in particular, generates enormous amounts of unstructured textual data across multiple domains, including clinical notes, discharge summaries, patient histories, research articles, and diagnostic reports. LLMs offer a potential solution by automatically processing and simplifying this information in a human-like manner, creating new possibilities for improving clinical workflows, patient engagement, and medical research. Studies by \cite{kung2023} and \cite{nori2023} showed that models like ChatGPT \cite{radford2018improving} and GPT-4 \cite{achiam2023gpt} can pass parts of the United States Medical Licensing Examination (USMLE), suggesting substantial biomedical knowledge acquisition. Moreover, domain-specific models such as ChatDoctor \cite{li2023chatdoctor} and Radiology-GPT \cite{liu2023radiology} show that fine-tuning general LLMs with specialized medical datasets can significantly enhance their contextual understanding and accuracy. The authors in \cite{ayers2023} further observed that patient-facing responses written by ChatGPT were often rated more empathetic and detailed than those from physicians on online forums, hinting at its potential role in patient education and telemedicine.

\begin{figure}[!b]
    \centering
    \includegraphics[width=0.8\linewidth]{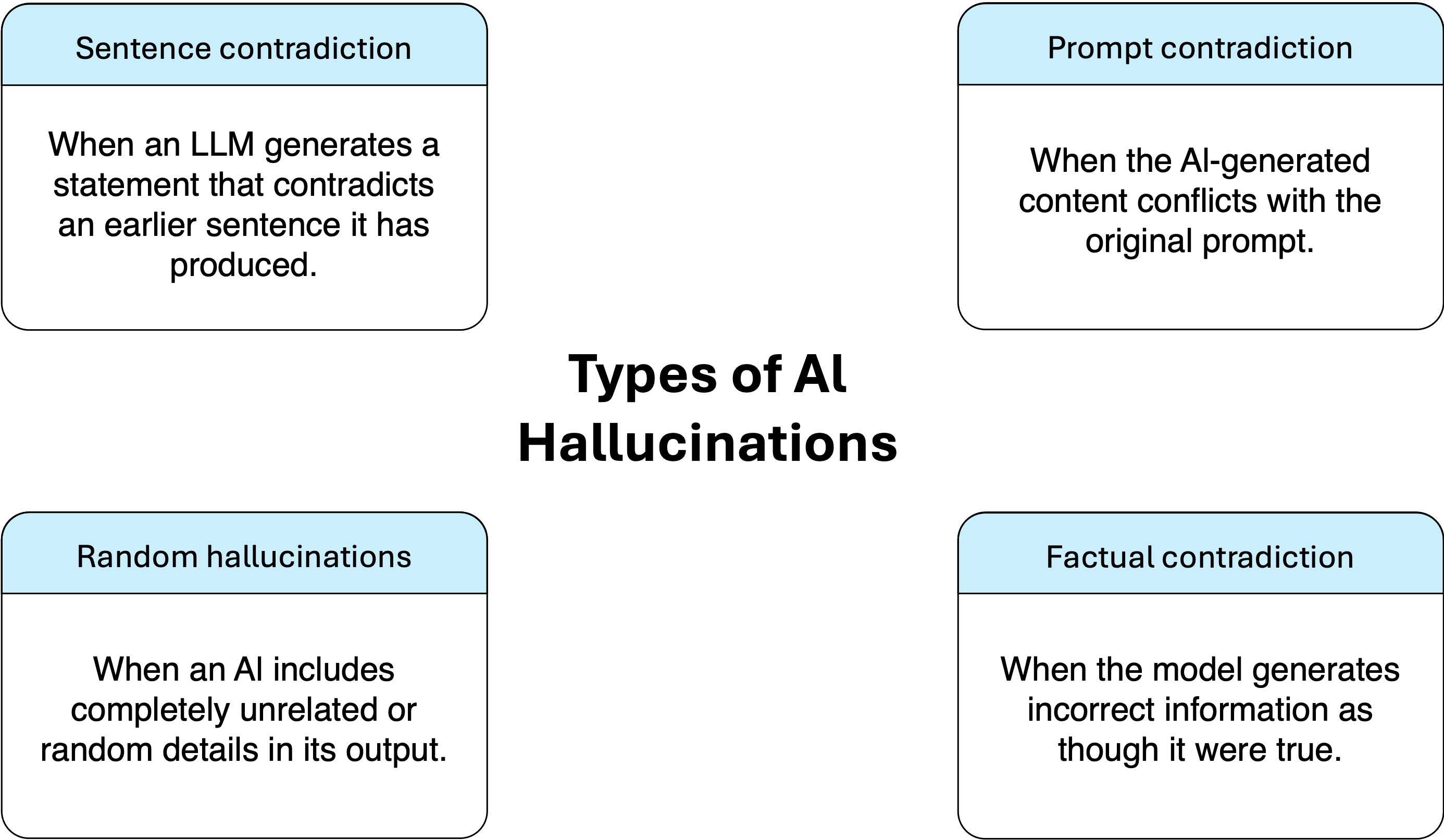}
    \caption{Types of Hallucinations in LLMs}
    \label{Types_of_AI_hallucinations}
     \vspace{-1\baselineskip}
\end{figure}

Despite this promise, the integration of LLMs into healthcare raises several critical challenges. A critical concern is accuracy and reliability in medical contexts. LLMs generate outputs probabilistically based on learned patterns rather than verified medical databases, which can lead to errors particularly problematic in healthcare settings. This issue is often referred to as \textit{hallucination}, where models produce fabricated or misleading information that sounds convincingly real. These hallucinations can be factual, where false statements are presented as facts; logical, involving internal contradictions within responses; or random, consisting of irrelevant content that appears contextually appropriate as shown in Figure \ref{Types_of_AI_hallucinations}. The authors in \cite{duong2024} observed that LLMs can generate confident but incorrect answers to genetics questions, illustrating the importance of evaluating model reliability across specialized medical domains. The authors in \cite{tam2024} emphasized that most evaluations depend on superficial text-similarity metrics like BLEU \cite{papineni2002bleu} or ROUGE \cite{lin2004rouge}, which reward lexical overlap rather than factual correctness, potentially causing the models to achieve high scores on standard benchmarks while producing clinically invalid or harmful outputs. These limitations highlight the need for rigorous, domain-specific evaluation frameworks before these tools can be trusted in sensitive healthcare environments.

Most existing studies focus on individual use cases or single models rather than benchmarking across multiple architectures and task types, creating a need for standardized comparative evaluation. LLMs are surveyed across medical specialties in \cite{mumtaz2024}, and it can be noted that nearly all evaluations were qualitative or task-specific, often limited to small datasets or synthetic prompts, focused on individual use cases rather than systematic comparison, and lacking in clinical validation. Similarly, \cite{yang2023healthcare} and \cite{wang2024} observed that while LLMs show promise in medical education and triage, few studies combine question answering and summarization tasks within one unified framework. LLMs are no longer research curiosities, they are rapidly entering patient-facing environments through chatbots and virtual health assistants, symptom checkers and triage tools, clinical-documentation assistants, Electronic Health Record (EHR) integration systems, and teleconsultation platforms. The studies in \cite{busch2025} and \cite{qin2025} called for empirical benchmarks to inform regulatory frameworks similar to those governing medical devices. This study provides empirical comparisons between general-purpose and domain-adapted LLMs to support evidence-based policymaking and deployment decisions. Hence, the contributions of this research are as follows.

\begin{enumerate}
\item A benchmarking framework is introduced that systematically compares five prominent LLMs across three complementary healthcare tasks (MedMCQA, PubMedQA, and Asclepius), using consistent prompt templates and a combined suite of quantitative metrics (Accuracy, Macro F1, BLEU, ROUGE-L, BERTScore) to enable direct, fair performance comparison.

\item The study demonstrates that LLM performance is architecture and task-dependent. Domain-tuned models (ChatDoctor) excel in semantic fidelity and clinical summarization, while general-purpose models (Grok 3 Mini) outperform on structured factual recall. This reveals fundamental trade-offs between discriminative and generative capabilities, supporting a task-routing deployment paradigm over universal model application in healthcare.

\item Our work exposes key limitations in standard evaluation metrics for medical AI, showing that different metrics capture fundamentally different performance dimensions and relying on any single metric risks an incomplete or misleading assessment of model behavior. We determined that BLEU/ROUGE reflect lexical fidelity while BERTScore captures semantic alignment.
\end{enumerate}

\begin{table*}[!t]
\centering
\small
\caption{Comparative Analysis of Related LLM Healthcare Evaluation Studies}
\renewcommand{\arraystretch}{1}
\begin{tabular}{|p{2cm}|p{5cm}|p{2cm}|p{3cm}|p{4cm}|}
\hline
\textbf{Study} & \textbf{Models Evaluated} & \textbf{Task Types} & \textbf{Datasets Used} & \textbf{Evaluation Metrics} \\
\hline
Kung et al. (2023) \cite{kung2023} & GPT-3.5 only & MCQ (USMLE) & USMLE Step 1-3 & Pass/Fail, Accuracy \\
\hline
Cascella et al. (2023) \cite{cascella2023evaluating} & ChatGPT only & Open-ended QA & Proprietary clinical prompts & Qualitative only \\
\hline
Singhal et al. (2023) \cite{singhal2023large} & Med-PaLM 2 vs GPT-4 & MCQ, long-form QA & MedQA, MedMCQA, HealthSearchQA & Accuracy, human ratings \\
\hline
Nori et al. (2023) \cite{nori2023capabilities} & GPT-4 only & MCQ & USMLE, MedQA & Accuracy only \\
\hline
Lievin et al. (2024) \cite{lievin2024can} & GPT-3/4, PaLM & MCQ reasoning & MedQA, MedMCQA & Accuracy, chain-of-thought \\
\hline
Van Veen et al. (2023) \cite{vanveen2023clinical} & 6 LLMs & Summarization only & Proprietary clinical notes & ROUGE, BERTScore \\
\hline
Jin et al. (2023) \cite{jin2023pubmedqa} & BERT variants & Binary QA & PubMedQA & Accuracy, F1 \\
\hline
Proposed Study & Grok 3 Mini, GPT-4o-Mini, Gemini 2.5 Flash Lite, LLaMA-3.1-8B, ChatDoctor & MCQ, Binary QA, Summarization & MedMCQA, PubMedQA, Asclepius & Accuracy, Macro F1, BLEU, ROUGE-L, BERTScore \\
\hline
\end{tabular}
\label{tab:comparison}
\end{table*}

\section{Literature Review}

Large language models (LLMs) have rapidly gained attention in healthcare due to their ability to process and generate complex medical text. However, while these models demonstrate impressive capabilities, their performance in high-stakes medical contexts raises important concerns regarding accuracy, safety, and ethical use. This review identifies methodological gaps in current evaluation practices and motivates the need for a standardized, multi-dimensional framework for assessing LLMs in healthcare.

The introduction of the transformer architecture by \cite{vaswani2017} marked a paradigm shift, enabling parallel processing and the self-attention mechanism that allows models to understand relationships between words across entire documents. This innovation led to the creation of Large Language Models, deep neural architectures trained on massive text corpora that can understand, reason, and generate human-like language. Examples include GPT-3, ChatGPT, LLaMA, Grok, and Gemini, which have demonstrated unprecedented generalisation capabilities across domains, including education, law, cybersecurity, and healthcare \cite{premalatha2025revolutionizing}. In healthcare, LLMs offer the ability to automate routine documentation, summarise long medical histories, extract clinical entities, and even generate drafts of discharge summaries, while supporting clinicians in evidence synthesis and helping patients understand their diagnoses. Large datasets such as PubMed \cite{canese2013pubmed}, MIMIC-III \cite{johnson2016mimic}, and MedMCQA \cite{pal2022medmcqa} can be leveraged to train or evaluate such models.

Several studies have demonstrated the versatility of LLMs in clinical communication and research. Patients rated ChatGPT responses as more empathetic and better structured than physician answers on online health forums \cite{ayers2023}, while ChatGPT was found to generate readable discharge summaries and research abstracts, reducing administrative workloads \cite{cascella2023evaluating}. The work in \cite{szabo2025comparative} evaluated multiple LLMs in medical education contexts and reported that they effectively explained histological concepts and improved learning outcomes. A parallel research line has focused on domain-specific fine-tuning: ChatDoctor \cite{li2023chatdoctor} and BioGPT \cite{peng2023} improve factual reliability while maintaining natural communication flow, and authors in \cite{algaradi2025} reinforced that model fine-tuning combined with Reinforcement Learning with Human Feedback (RLHF) remains the most effective way to reduce hallucination frequency. Yet even these refined systems face challenges in explainability, as the mechanisms behind LLM reasoning remain largely opaque.

The study in \cite{busch2025} surveyed over 150 publications and found that while 72\% demonstrated positive results, less than 20\% employed reproducible datasets such as MedMCQA or PubMedQA \cite{agrawal2022,jin2019pubmedqa}, and fewer still combined multiple metrics in a unified evaluation scheme. Another work \cite{tam2024} proposed that without consistent benchmarks, even high-performing models may give an illusion of reliability that does not hold under real clinical scrutiny. A concise comparison between previous studies and this work is provided in Table \ref{tab:comparison}.

The review of literature across more than twenty-five studies reveals three recurring patterns. First, general-purpose LLMs such as ChatGPT and Grok demonstrate impressive versatility and linguistic performance but often lack domain-specific factual grounding. Second, fine-tuned models like ChatDoctor, Radiology-GPT, and Med-PaLM achieve improved contextual accuracy but still face issues of transparency and dataset bias. Third, most prior evaluations test single models on specific datasets using varied metrics, preventing adequate cross-model comparison and limiting insights into task-dependent performance patterns. This study addresses that gap by benchmarking multiple LLMs across standardized datasets representing different clinical tasks, using reproducible metrics to reveal performance patterns that can inform appropriate model selection for specific healthcare applications.

\begin{table*}[!t]
\centering
\normalsize
\caption{Datasets and task mapping.}
\label{tab:datasets}
\begin{tabular}{|p{4cm}|p{3cm}|p{3cm}|p{5cm}|}
\hline
\textbf{} & \textbf{MedMCQA} & \textbf{PubMedQA} & \textbf{Asclepius} \\
\hline
\textbf{Type} & MCQ & Yes/No/Maybe & Clinical text \\
\hline
\textbf{Task} & Question Answers & Question Answers & Summarization \\
\hline
\textbf{Size} & 183k questions & 1k labeled & 2k notes \\
\hline
\textbf{Primary Metrics} & Accuracy, Macro~F1 & Accuracy, Macro~F1 & BLEU, ROUGE-L, BERTScore \\
\hline
\end{tabular}
\end{table*}
\section{Research Methodology}
\label{sec:methodology}

\begin{figure}[t]
    \centering
    \includegraphics[width=1\linewidth]{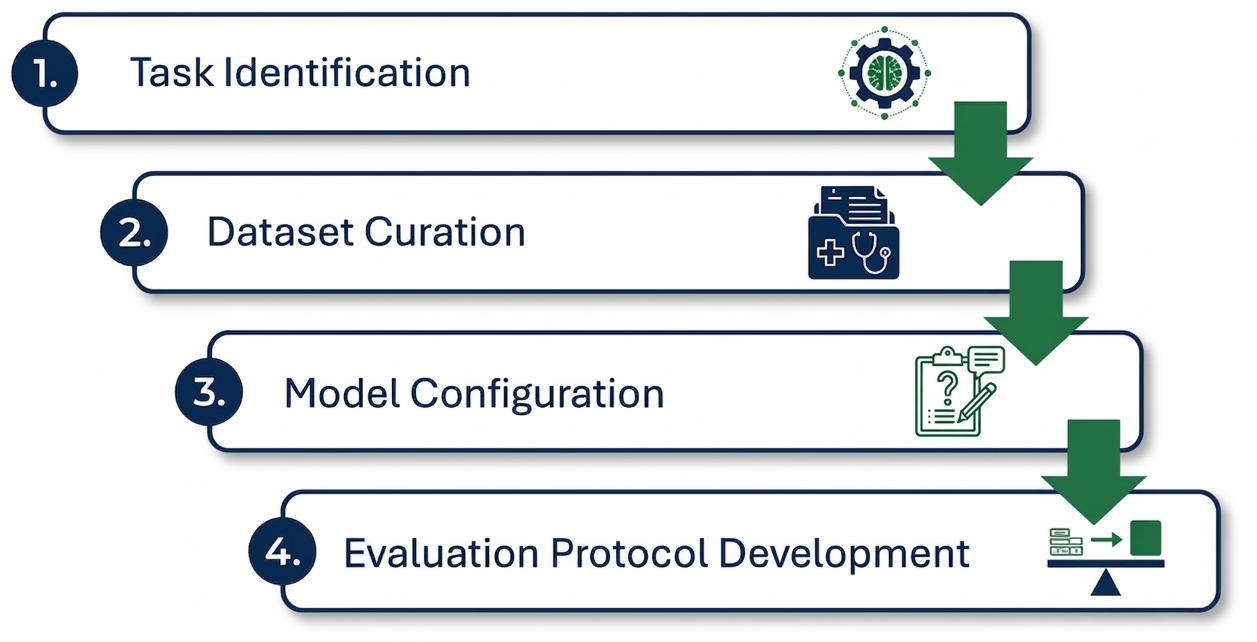}
    \caption{Methodology Workflow.}
    \label{Methodology_Workflow_Diagram}
     \vspace{-1\baselineskip}
\end{figure}

This research adopts a structured, multi-phase methodology designed to evaluate the performance, safety, and reliability of large language models (LLMs) within the healthcare domain. Models were selected to represent: (1) domain-specific medical fine-tuning (ChatDoctor), (2) state-of-the-art general-purpose systems (GPT-4o-Mini, Grok 3 Mini, Gemini), and (3) open-source reproducibility (LLaMA-3.1-8B). The design follows four major phases discussed in subsequent sections, with Figure \ref{Methodology_Workflow_Diagram} presenting a complete end-to-end overview of the experimental pipeline.

\subsection{Phase 1: Task Identification}
After reviewing over 25 recent studies in medical NLP, three core task categories were identified: medical multiple-choice question answering (MCQ QA), medical Yes/No/Maybe question answering (YNM QA), and clinical note summarization. MCQ questions, sourced from MedMCQA \cite{pal2022medmcqa}, test factual recall, diagnostic reasoning, and terminological precision, exposing whether models have genuine biomedical knowledge rather than just linguistic fluency. YNM questions from PubMedQA \cite{canese2013pubmed} require the model to read a research abstract and determine whether evidence supports, contradicts, or remains inconclusive about the question, assessing nuanced reasoning that reflects real medical judgment. The Asclepius Synthetic Clinical Notes dataset \cite{kweon2024publicly} supports the summarization task, evaluating whether LLMs can generate coherent, factual, and medically safe summaries for clinical decision support and documentation automation. Together, these three tasks provide a balanced evaluation framework spanning discriminative and generative medical capabilities.

\subsection{Phase 2: Dataset Curation}
This study relies exclusively on publicly available, non-identifiable datasets that pose no risk to patient privacy. Three datasets, MedMCQA \cite{pal2022medmcqa}, PubMedQA \cite{canese2013pubmed}, and Asclepius Synthetic Clinical Notes \cite{kweon2024publicly}, were selected based on coverage, quality, and relevance to clinical tasks, as summarized in Table \ref{tab:datasets}.

\textbf{MedMCQA:} Developed by \cite{lobo2025impact}, this dataset contains more than 183,000 multiple-choice questions spanning 21 subjects including anatomy, pharmacology, pathology, and community medicine, collected from Indian medical entrance examinations (NEET-PG and AIIMS). It tests both factual recall and conceptual reasoning, and each question is categorized by subject and difficulty level, enabling detailed performance comparison across subdomains and verification of whether a model generalizes medical knowledge rather than relying on pattern recognition.

\textbf{PubMedQA:} Derived from PubMed \cite{jin2019pubmedqa}, this dataset consists of approximately 1,000 labeled question-answer pairs where each answer is annotated as `Yes', `No', or `Maybe', reflecting whether research findings support, contradict, or remain inconclusive about the question. Unlike traditional question answering, this dataset tests whether models can recognize ambiguity, acknowledge uncertainty, and respond conservatively when evidence is unclear. This is crucial for healthcare, where overconfident or incorrect assertions can be unsafe.

\textbf{Asclepius:} The Asclepius Synthetic Clinical Notes dataset \cite{kweon2024publicly} consists of synthetically generated medical records closely mimicking real clinical documentation, each accompanied by a human-written gold-standard summary. It provides an ethically safe and scalable alternative to hospital EHRs, allowing evaluation of summarization systems without violating data protection regulations. High performance on Asclepius indicates that a model can potentially assist in documentation tasks such as discharge summaries or progress note generation while maintaining clarity and safety in medical communication.

All datasets were uniformly preprocessed through text normalization, metadata removal, and verification of encoding consistency to minimize bias and ensure comparable tokenization across models.

\subsection{Phase 3: Model Configuration}

\textbf{ChatDoctor:} ChatDoctor is a domain-specific LLM fine-tuned from Meta's LLaMA architecture using a diverse corpus of medical dialogues, clinical question-answer pairs, and health-related case discussions integrating PubMed abstracts, MedQA, medical textbooks, and curated doctor-patient interactions. A distinguishing feature is its built-in safety alignment layer, which penalizes unsafe or speculative outputs, allowing the model to express uncertainty using controlled language while recommending further clinician evaluation. It follows a human-in-the-loop paradigm, supporting clinicians by reducing repetitive tasks while leaving the final decision to human expertise, and serves in this study as a benchmark for domain-tuned LLMs.

\textbf{Grok 3 Mini:} Grok 3 Mini is a lightweight general-purpose reasoning model trained on a vast multi-domain corpus including encyclopedic sources, technical manuals, and internet-scale datasets, optimized for reduced inference latency. Despite lacking domain-specific fine-tuning, it displays impressive generalization across reasoning tasks due to its reinforcement-learning-based optimization and a curriculum that prioritizes logical coherence and factual grounding. Its inclusion provides an empirical reference point for measuring the advantages of domain fine-tuning, revealing the trade-off between adaptability and medical caution.

\textbf{GPT-4o-Mini:} GPT-4o-Mini is a compact version of OpenAI's GPT-4 Omni architecture, evaluated here in text-only mode, incorporating RLHF and safety-layer moderation. It demonstrates strong reasoning ability and linguistic coherence and functions in this study as a high-quality general baseline to measure how much value domain adaptation adds beyond a strong general model. Its advanced token-compression techniques reduce computational load while preserving contextual reasoning depth, making it ideal for large-scale batch testing.

\textbf{Gemini 2.5 Flash Lite:} Part of Google DeepMind's Gemini family, this lightweight high-throughput model is engineered for real-time summarization and information retrieval, employing a dual-pass decoding mechanism that first generates a concise outline and then refines it for clarity and consistency. It lacks the deep domain fine-tuning of ChatDoctor, making it less cautious in speculative contexts, but its low inference time and stable summarization quality make it a practical choice for scenarios like automated reporting or patient-record summarization.

\textbf{LLaMA-3.1-8B:} Developed by Meta AI, this open-source model with approximately 8 billion parameters serves two purposes: as a baseline for evaluating the benefits of fine-tuning (by comparing its outputs with ChatDoctor) and as an example of an open model capable of competitive reasoning without proprietary constraints. Its transparency, full access to its architecture, weights, and training documentation, makes it a preferred choice for academic and reproducible experiments and ensures that all experiments can be replicated, reviewed, and extended by future researchers.

To ensure fairness, each model received identical or semantically equivalent prompts. All model outputs were stored in a unified JSON structure including the input, output, model name, timestamp, and evaluation metrics, ensuring consistent downstream analysis.

\subsection{Phase 4: Evaluation Protocol Development}
Evaluation is the most critical part of this methodology because it determines not only performance but also safety, coherence, and factual correctness. To develop the evaluation protocol, we employ the following metrics:

\begin{itemize}[leftmargin=0pt,label={}]

\item \vspace{0.2cm}\textbf{Accuracy and Macro F1 (QA):} For question-answering tasks, we employ accuracy to measure the proportion of correct responses and Macro F1 \cite{sokolova2009systematic} to assess balanced performance across answer classes, averaging class-wise F1 scores and giving equal weight to all categories regardless of frequency:
\begin{align}
\mathrm{Macro\ F1} &= \frac{1}{C}\sum_{i=1}^C \frac{2 \cdot \mathrm{Precision}_i \cdot \mathrm{Recall}_i}{\mathrm{Precision}_i + \mathrm{Recall}_i}.
\end{align}
This dual-metric approach is essential in healthcare contexts where class imbalance is common (e.g., rare but critical ``Maybe'' responses in PubMedQA), ensuring models demonstrate both correctness and balanced coverage rather than exploiting dataset biases.

\item \vspace{0.2cm}\textbf{BLEU (Summarization):} BLEU \cite{papineni2002bleu} measures n-gram overlap between generated and reference summaries, quantifying lexical precision and grammatical fluency:
\begin{equation}
\mathrm{BLEU} = \mathrm{BP}\cdot \exp \left(\sum_{n=1}^{N} w_n \log p_n \right),
\end{equation}
where $p_n$ represents modified n-gram precision and $\mathrm{BP}$ is a brevity penalty. While BLEU effectively captures surface-level linguistic fidelity, it penalizes semantically equivalent paraphrases, necessitating complementary semantic metrics.

\item \vspace{0.2cm}\textbf{ROUGE-L (Summarization):} ROUGE-L \cite{lin2004rouge} evaluates content coverage by computing the longest common subsequence (LCS) between generated and reference texts:
\begin{equation}
\mathrm{ROUGE\mbox{-}L} = \frac{\mathrm{LCS}(\hat{y},y)}{|y|}.
\end{equation}
By emphasizing recall over precision, ROUGE-L rewards summaries that preserve critical clinical information such as diagnoses, treatments, and outcomes, making it particularly suitable for medical summarization where completeness is paramount.

\item \vspace{0.2cm}\textbf{BERTScore (Summarization):} BERTScore \cite{zhang2020bertscore} assesses semantic similarity using contextual embeddings from pre-trained BERT models:
\begin{equation}
\mathrm{BERTScore\ F1} = \frac{2 \cdot P_{\mathrm{bert}} \cdot R_{\mathrm{bert}}}{P_{\mathrm{bert}} + R_{\mathrm{bert}}}.
\end{equation}
Unlike n-gram metrics, BERTScore recognizes semantically equivalent expressions (e.g., ``type 2 diabetes'' and ``non-insulin dependent diabetes''), providing a robust measure of meaning preservation essential for evaluating clinical text where lexical variation is common but semantic accuracy is critical.

\end{itemize}

This multi-metric evaluation framework ensures comprehensive assessment: BLEU and ROUGE-L capture linguistic structure and content coverage, while BERTScore validates semantic fidelity. Together these metrics enable distinguishing models that achieve surface-level similarity from those that preserve clinical meaning.

\section{Experimental Evaluation}

The experiments were conducted using a hybrid setup where local execution for orchestration and data processing is performed, and cloud-based inference for large-scale model testing is utilized. Hardware included a MacBook Pro (M4, 24 GB RAM) for open-source models (LLaMA-3.1-8B and ChatDoctor), while proprietary models (GPT-4o-Mini, Gemini 2.5 Flash Lite, and Grok 3 Mini) were accessed via official APIs to ensure version stability. Python 3.10 was the primary programming language; random seeds were fixed to maintain consistency in sampling, and logs were automatically generated recording timestamps, model versions, temperature settings, and output lengths. This controlled setup ensured that differences in results were attributed to model behavior rather than random environmental or implementation factors.

Accuracy and Macro F1 are used jointly for question-answering tasks to differentiate raw correctness from balanced decision-making, particularly in the presence of uncertainty and class imbalance, discrepancies between them are treated as indicators of overconfident or biased response tendencies. For clinical summarization, BERTScore serves as the primary indicator of semantic fidelity, with BLEU and ROUGE-L providing complementary views of linguistic structure and informational coverage. Strong performance is defined not by peak scores on a single metric, but by consistent alignment across factual accuracy, semantic integrity, and cautious language generation.

\begin{figure}[t]
    \centering
    \begin{minipage}{0.35\textwidth}
        \centering
        \includegraphics[width=\linewidth]{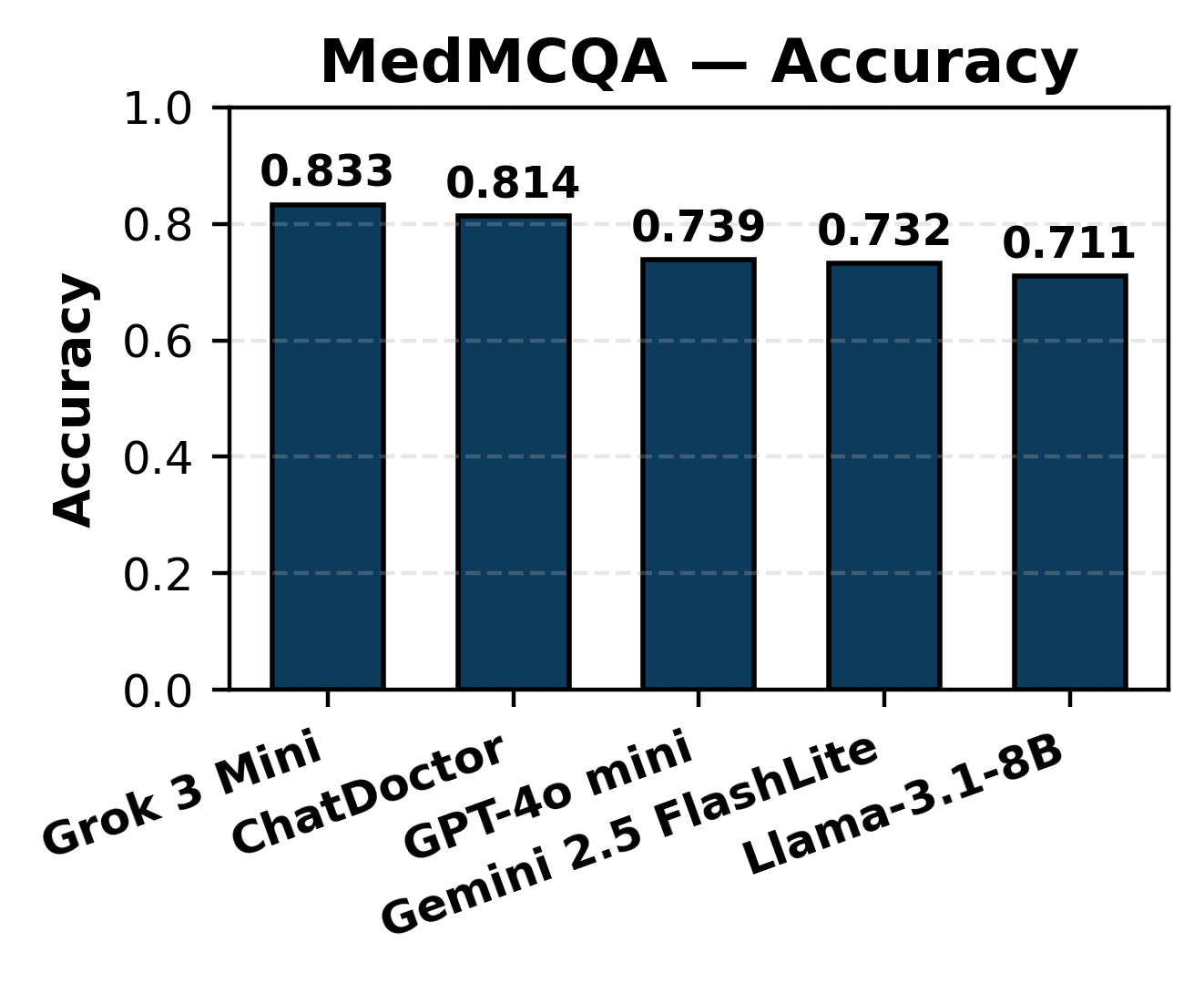}
        \caption{MEDMCQA - Accuracy}
        \label{fig:medmcqa_accuracy}
    \end{minipage}
    % \hfill
    \begin{minipage}{0.35\textwidth}
        \centering
        \includegraphics[width=\linewidth]{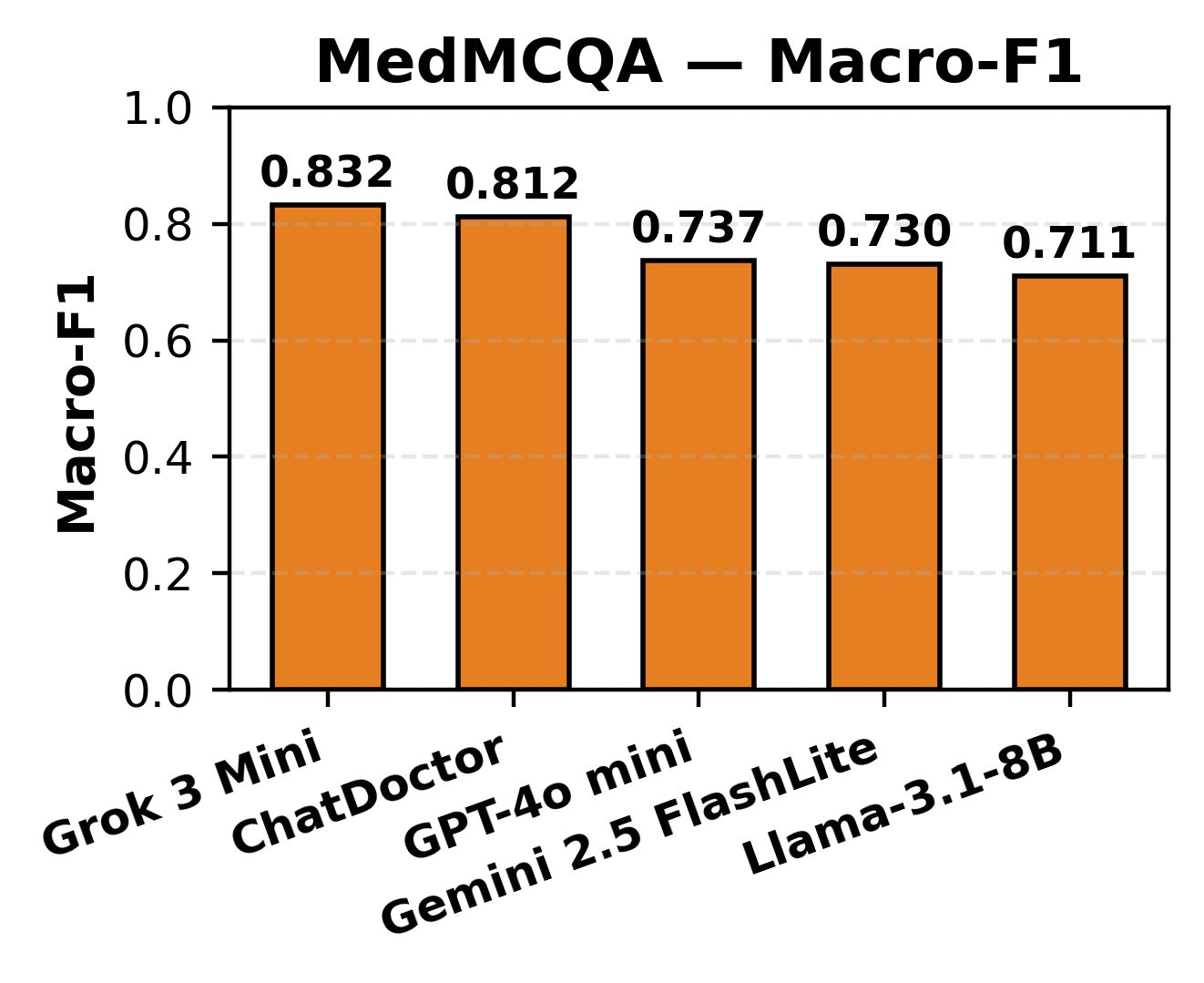}
        \caption{MEDMCQA - F1}
        \label{fig:medmcqa_f1}
    \end{minipage}
    \vspace{-1\baselineskip}
\end{figure}

\subsection{Experimental Case Studies and Results}
This section presents the quantitative findings from three experimental case studies
designed to evaluate complementary dimensions of healthcare-oriented natural language processing: (i) factual question answering, (ii) reasoning under uncertainty,
and (iii) clinical summarization. These tasks examine both knowledge retrieval and
text generation capabilities of large language models (LLMs) in medical contexts. All
experiments employed standardized prompt formats, identical evaluation conditions,
and quantitative metrics, ensuring fair cross-model comparison.

\textbf{Case Study 1: Medical MCQ Answering}

The models were evaluated on the MedMCQA dataset through a unified prompt template:
\begin{tcolorbox}[colback=white!80!gray,colframe=white!80!gray]
{\fontfamily{qcr}\selectfont `Answer the following medical question accurately. Choose the correct option (A, B, C or D).'}
\end{tcolorbox}
This task tested factual recall, diagnostic reasoning, and terminological precision under identical test conditions.

Accuracy values ranged between 0.711 and 0.833 (Figure \ref{fig:medmcqa_accuracy}), with Macro-F1 scores between 0.711 and 0.832 (Figure \ref{fig:medmcqa_f1}), indicating consistent performance across different model architectures. High F1 values signify that models maintained both strong precision and recall across a diverse range of question types, confirming that outputs are grounded in retained medical information rather than random guessing.

\begin{figure}[H]
    \centering
    \begin{minipage}{0.35\textwidth}
        \centering
        \includegraphics[width=\linewidth]{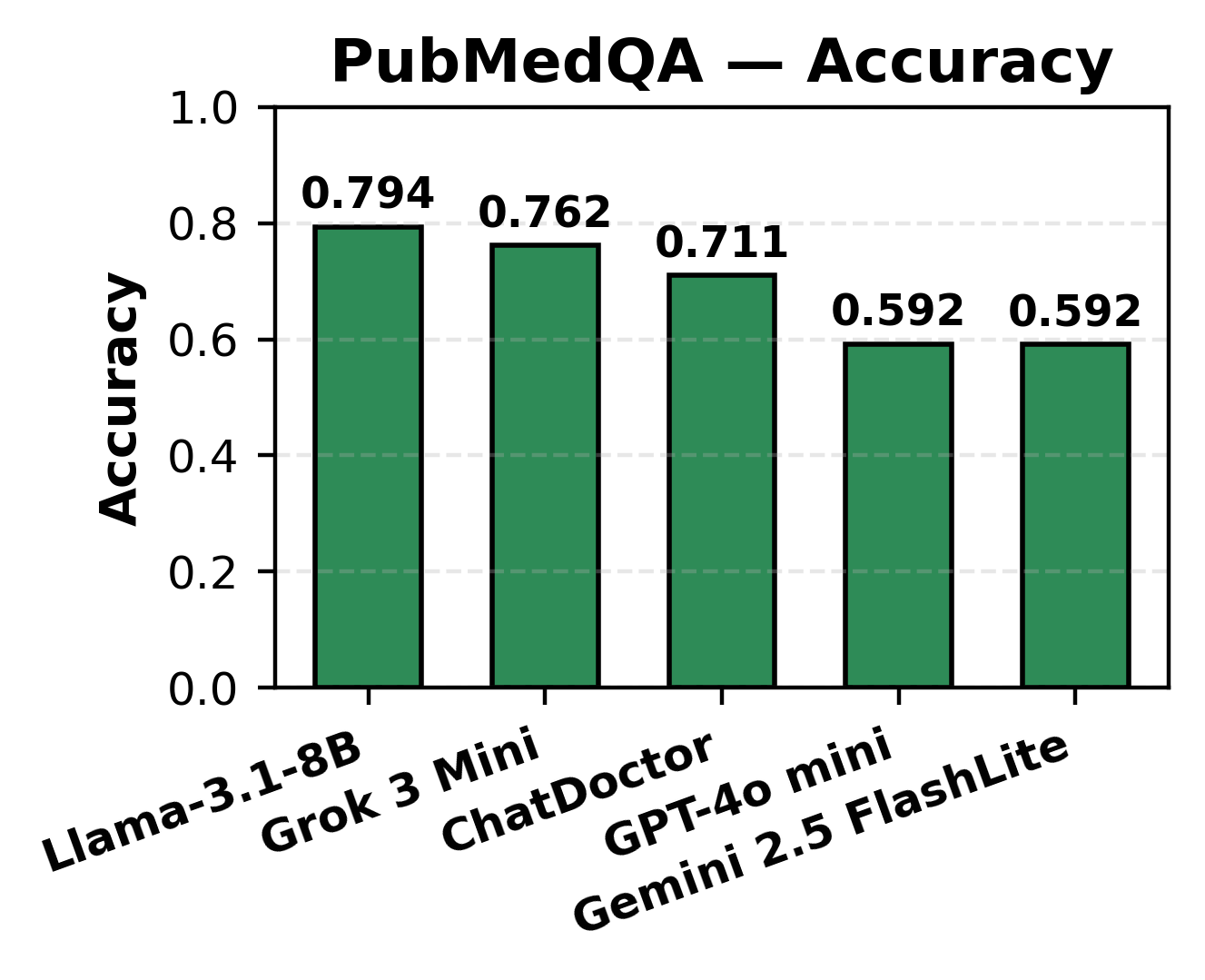}
        \caption{PubMedQA - Accuracy}
        \label{fig:pubmed_accuracy}
    \end{minipage}
    % \hfill
    \begin{minipage}{0.35\textwidth}
        \centering
        \includegraphics[width=\linewidth]{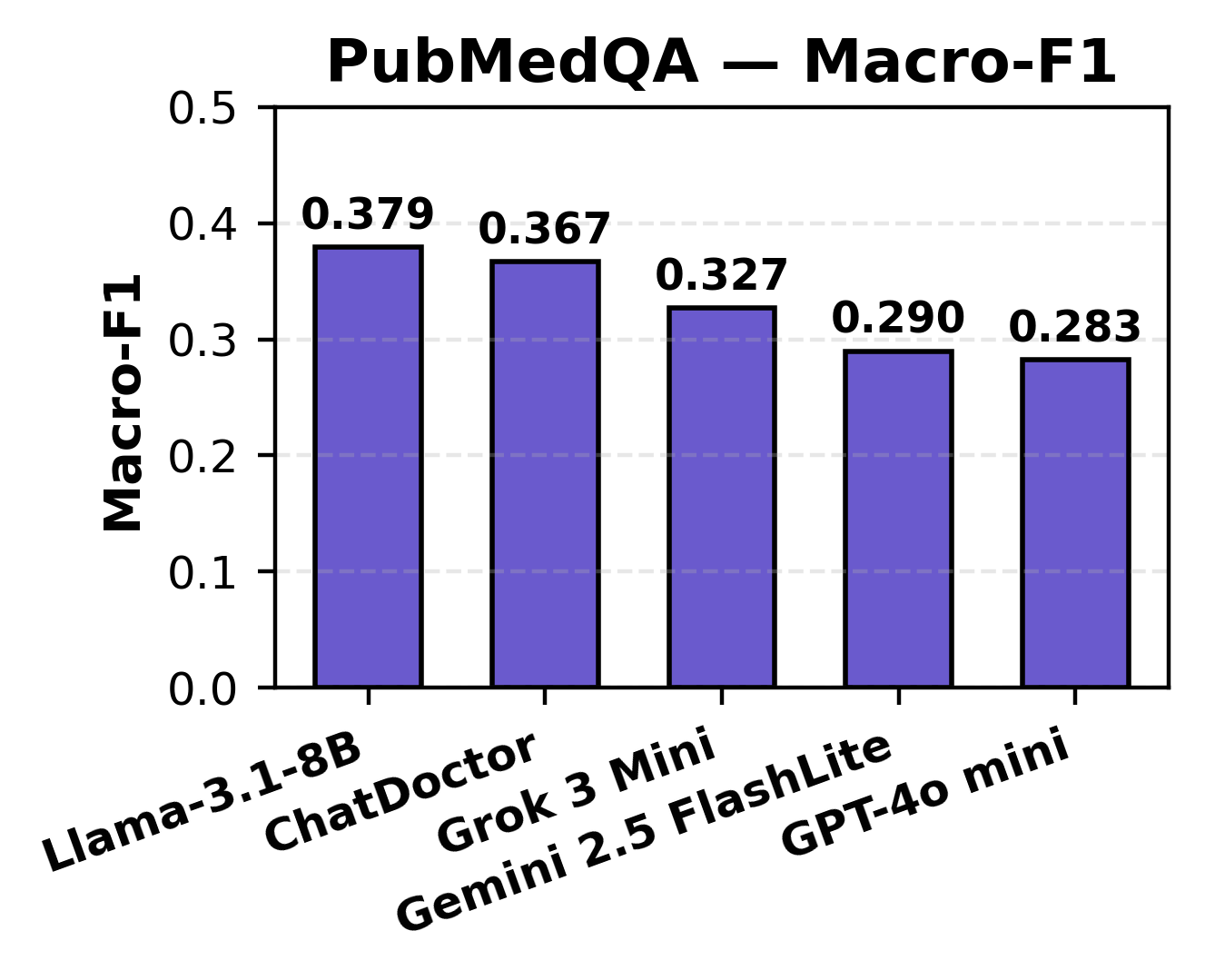}
        \caption{PubMedQA - F1}
        \label{fig:pubmed_f1}
    \end{minipage}
    \vspace{-1\baselineskip}
\end{figure}

\begin{tcolorbox}[colback=white!80!gray,colframe=white!20!gray]
All evaluated models performed consistently well on medical multiple-choice questions, achieving high Accuracy and Macro-F1 scores ($\sim 0.71$-$0.83$). This indicates that, under a standardized prompt, the models can reliably recall and apply structured medical knowledge across both basic and clinical topics, demonstrating stable exam-style reasoning rather than random guessing or superficial pattern matching.
\end{tcolorbox}

% Source - https://tex.stackexchange.com/a/247935
% Posted by hadi
% Retrieved 2026-04-12, License - CC BY-SA 3.0

\textbf{Case Study 2: Biomedical Yes/No/Maybe Reasoning}

The PubMedQA dataset was used to assess reasoning under uncertainty, with each model receiving a biomedical research question and its corresponding abstract. The prompt instructed: 
\begin{tcolorbox}[colback=white!80!gray,colframe=white!80!gray]
{\fontfamily{qcr}\selectfont `Based on the research abstract below, answer Yes, No or Maybe depending on whether the findings support the question.'}
\end{tcolorbox}
This case study evaluates how effectively models can interpret scientific evidence and make cautious, evidence-based judgments which are crucial for clinical decision support systems.

Accuracy scores ranged from 0.592 to 0.794 (Figure \ref{fig:pubmed_accuracy}), confirming that models successfully interpreted biomedical text to derive logical answers from evidence-based abstracts. Macro-F1 scores ranged between 0.283 and 0.379 (Figure \ref{fig:pubmed_f1}), showing moderate but consistent ability to maintain balance across all three output classes. The `Maybe' category is rare but medically important; a model's ability to correctly identify uncertainty indicates higher clinical awareness and responsible interpretive behavior.

\begin{tcolorbox}[colback=white!80!gray,colframe=white!20!gray]
While models showed reasonable accuracy in interpreting biomedical abstracts ($\sim 0.59$-$0.79$), their substantially lower Macro-F1 scores indicate that reasoning under uncertainty remains challenging. In particular, correctly balancing Yes, No, and especially the clinically critical Maybe class is harder than factual recall, highlighting an important limitation for cautious clinical decision support.
\end{tcolorbox}

\textbf{Case Study 3: Clinical Note Summarization}

Models were evaluated on the Asclepius Synthetic Clinical Notes dataset with the prompt: 
\begin{tcolorbox}[colback=white!80!gray,colframe=white!80!gray]
{\fontfamily{qcr}\selectfont `Summarize the following clinical note in two sentences focusing on diagnosis and treatment outcome. Avoid adding information not present in the text.'}
\end{tcolorbox}
This task evaluated the models' ability to produce concise, coherent, and factually consistent summaries while retaining all clinically significant details.

\begin{figure}[!t]
    \centering
    \begin{minipage}{0.32\textwidth}
        \centering
        \includegraphics[width=\linewidth]{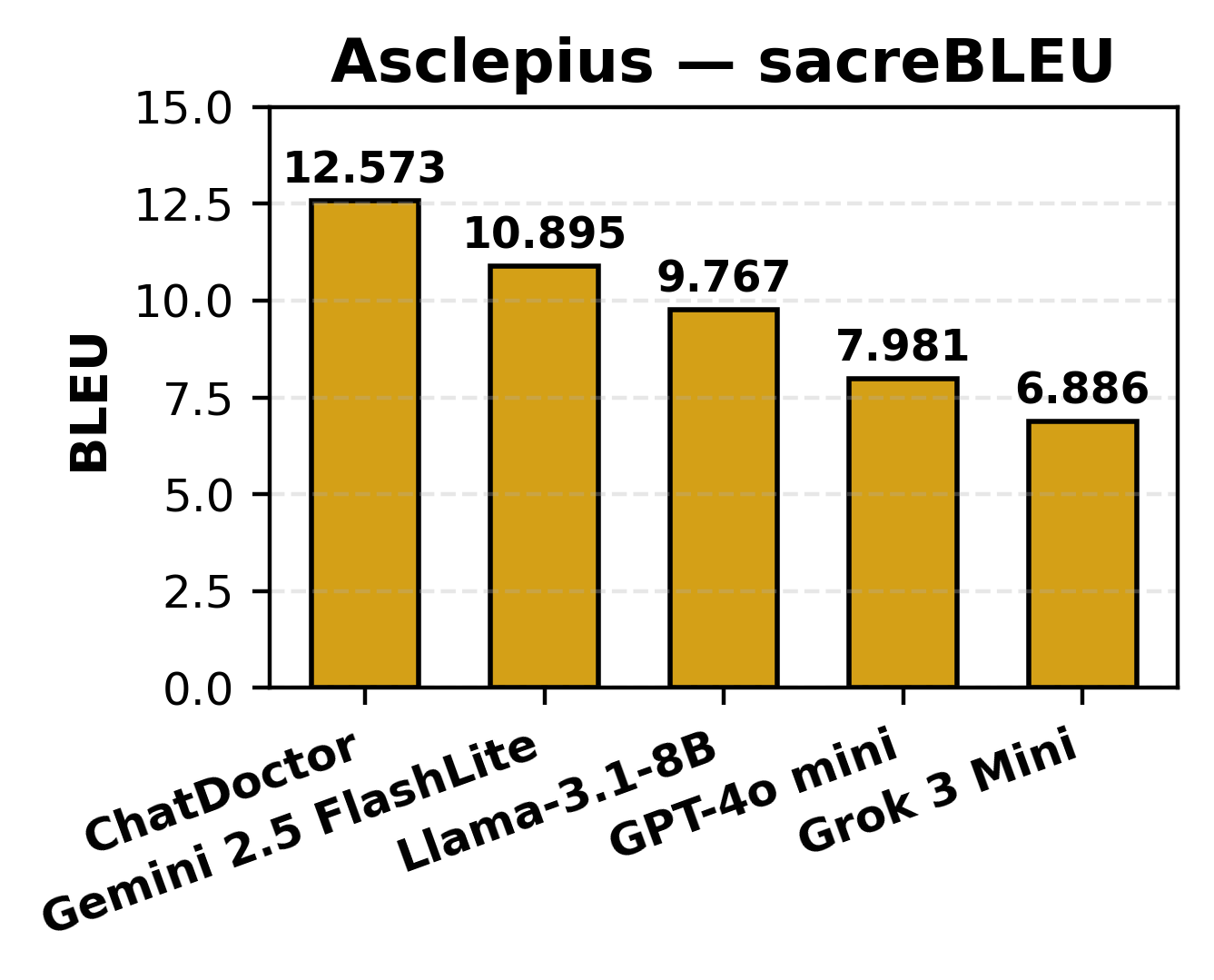}
        \caption{Asclepius - BLEU}
        \label{fig:asc_bleu}
    \end{minipage}
    \hfill
    \begin{minipage}{0.32\textwidth}
        \centering
        \includegraphics[width=\linewidth]{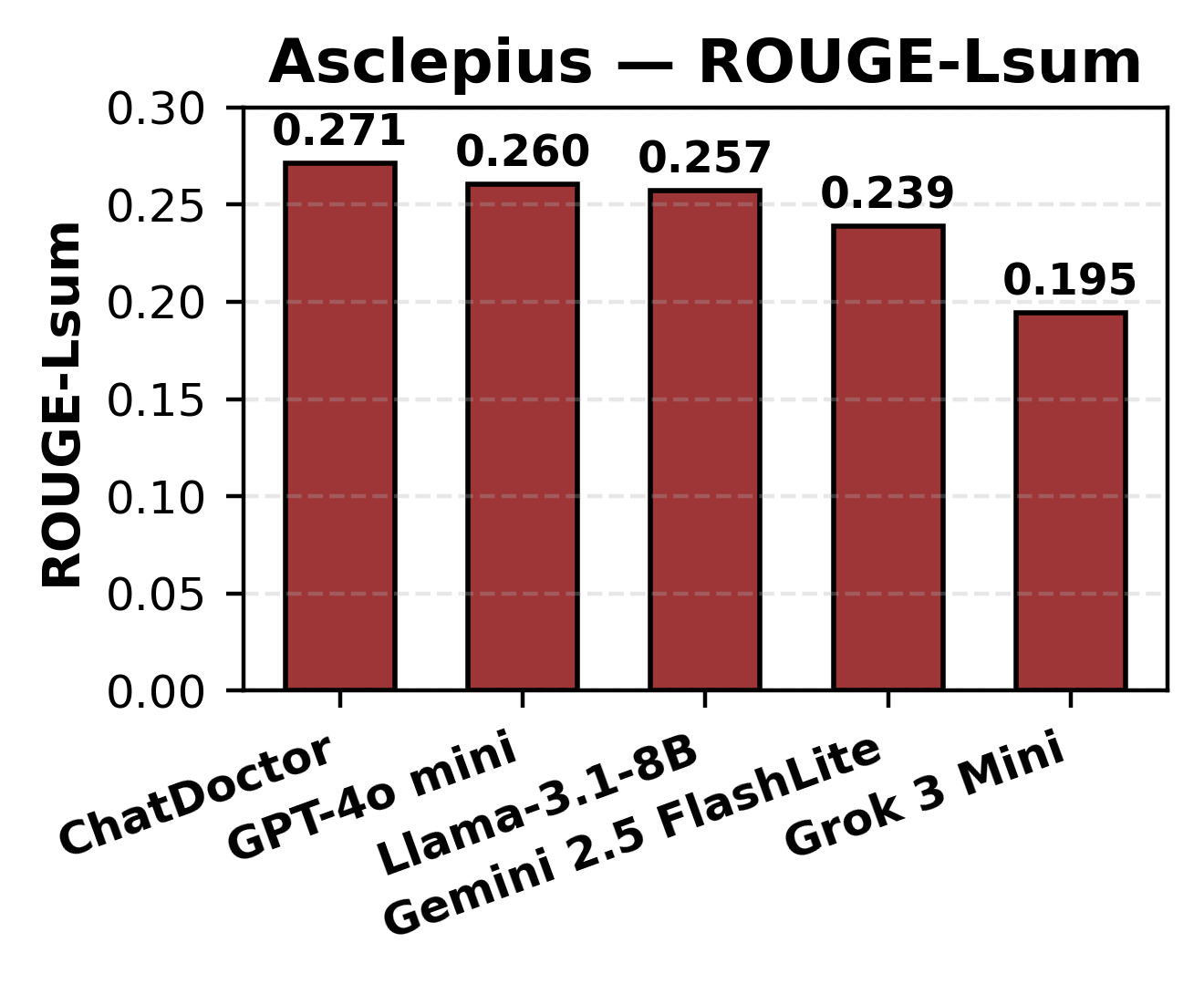}
        \caption{Asclepius - ROUGE-L}
        \label{fig:asc_rouge}
    \end{minipage}
    \hfill
    \begin{minipage}{0.32\textwidth}
        \centering
        \includegraphics[width=\linewidth]{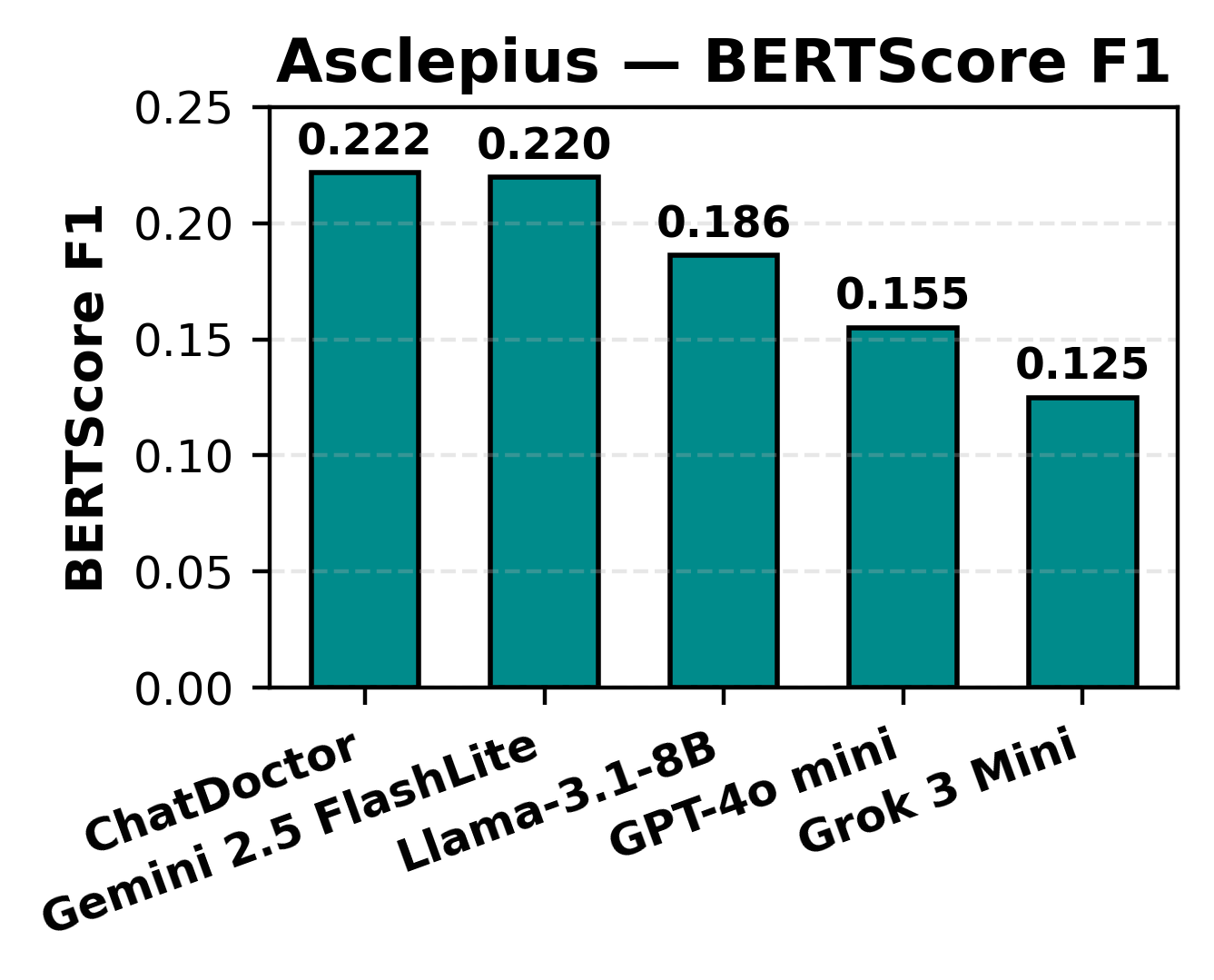}
        \caption{Asclepius - BERTScore F1}
        \label{fig:asc_bert}
    \end{minipage}
    \vspace{-1\baselineskip}
\end{figure}

BLEU values ranged from 6.886 to 12.573 (Figure \ref{fig:asc_bleu}), confirming linguistic fluency and precision. ROUGE-L scores ranged between 0.195 and 0.271 (Figure \ref{fig:asc_rouge}), reflecting adequate retention of essential medical facts including diagnosis, treatments, and outcomes. BERTScore F1 results, ranging from 0.125 to 0.222 (Figure \ref{fig:asc_bert}), show how semantically close the generated summaries were to the ground-truth references, capturing meaning preservation even when sentence structures differed.

\begin{tcolorbox}[colback=white!80!gray,colframe=white!20!gray]
All evaluated models were able to generate concise and medically coherent summaries from long clinical notes, preserving core diagnostic and treatment information without introducing hallucinations. Performance differences reveal that some models (e.g., ChatDoctor, Gemini) prioritize semantic fidelity and clinical safety over surface-level lexical overlap, highlighting their suitability for real-world documentation tasks such as discharge summaries and progress notes.
\end{tcolorbox}

\section{Discussion}

\begin{figure*}[!ht]
    \centering
    \includegraphics[trim={0.1cm 0.1cm 0.1cm 0cm},clip,width=1.0\textwidth]{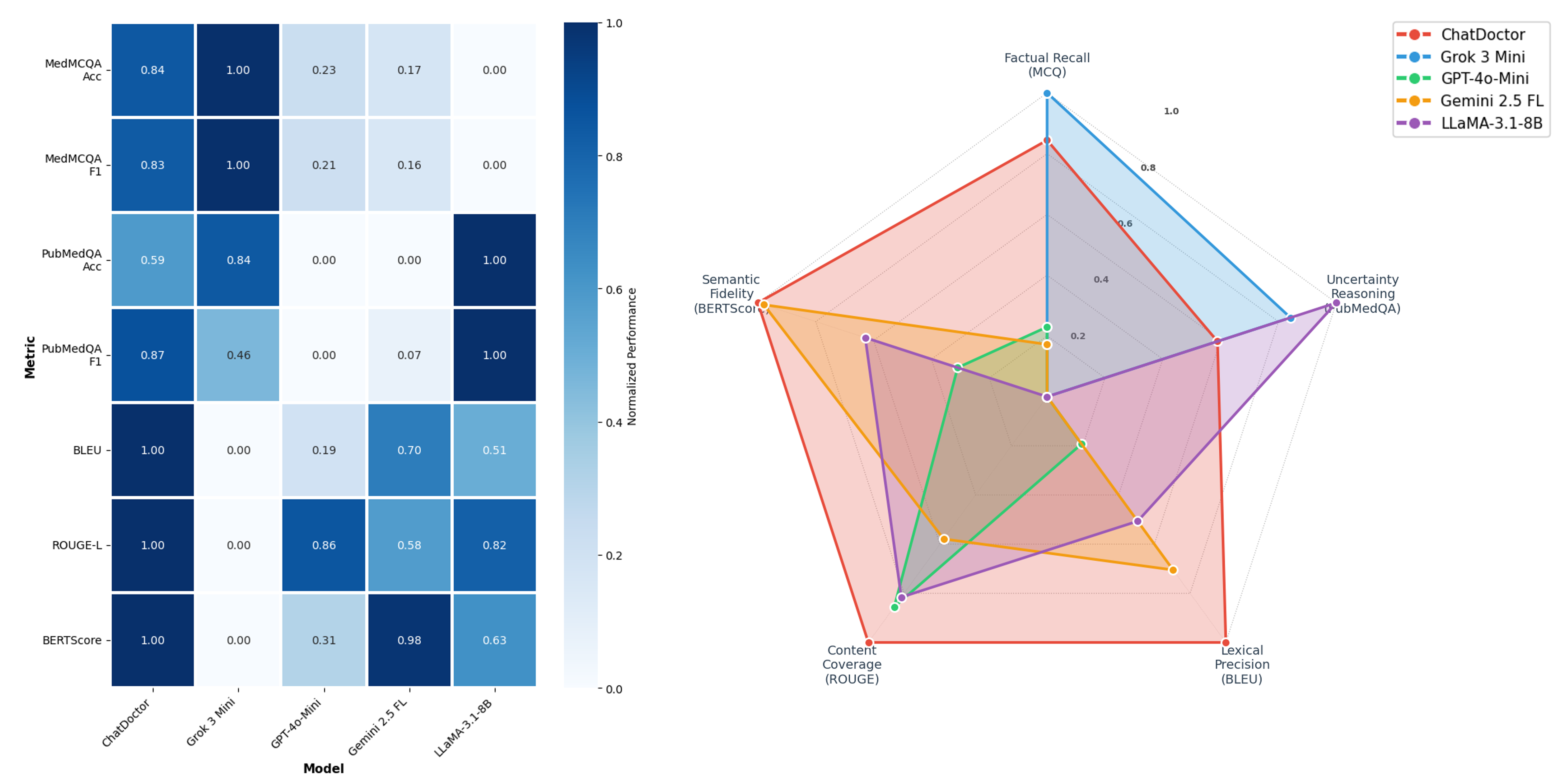}
    \caption{(Left) Normalized performance heatmap showing scaled scores. (Right) Radar plot displaying multi-dimensional capabilities of LLMs.}
    \label{fig:heatmap}
     \vspace{-1\baselineskip}
\end{figure*}

This study reveals substantial variation in how large language models perform across different healthcare tasks. Table~\ref{tab:results} and Figure~\ref{fig:heatmap} provide a consolidated view of model performance across factual recall (MedMCQA), reasoning under uncertainty (PubMedQA), and clinical summarization (Asclepius). The results reveal pronounced task-dependent specialization rather than uniform excellence. 

\begin{table}[H]
\centering
\caption{Comprehensive Results of LLM Performance Across All Experimental Scenarios}
\tinyscript
\setlength{\tabcolsep}{1.5pt}
\renewcommand{\arraystretch}{2}
\begin{tabular}{|l|c|c|c|c|c|c|c|}
\hline
\textbf{Model} &
\makecell{\text{MedMCQA} \\ (Acc)} &
\makecell{\text{MedMCQA} \\ (F1)} &
\makecell{\text{PubMedQA} \\ (Acc)} &
\makecell{\text{PubMedQA} \\ (F1)} &
\makecell{\text{Asclepius} \\ (BLEU)} &
\makecell{\text{Asclepius} \\ (ROUGE-L)} &
\makecell{\text{Asclepius} \\ (BERT F1)} \\
\hline
\textbf{ChatDoctor} & 0.814 & 0.812 & 0.711 & 0.367 & \textbf{12.573} & \textbf{0.271} & \textbf{0.222} \\
\hline
\textbf{Grok} & \textbf{0.833} & \textbf{0.832} & 0.762 & 0.327 & 6.886 & 0.195 & 0.125 \\
\hline
\textbf{GPT-4o} & 0.739 & 0.737 & 0.592 & 0.283 & 7.981 & 0.260 & 0.155 \\
\hline
\textbf{Gemini} & 0.732 & 0.730 & 0.592 & 0.290 & 10.895 & 0.239 & 0.220 \\
\hline
\textbf{LLaMA} & 0.711 & 0.711 & \textbf{0.794} & \textbf{0.379} & 9.767 & 0.257 & 0.186 \\
\hline
\end{tabular}
\label{tab:results}
\end{table}
Grok~3~Mini achieves the strongest performance in structured factual recall but shows limited semantic fidelity in summarization, whereas ChatDoctor exhibits the inverse pattern, excelling in clinically faithful text generation while maintaining moderate question-answering accuracy. LLaMA-3.1-8B stands out in uncertainty reasoning on PubMedQA despite weaker MCQ performance, highlighting its strength in evidence-based interpretation.

The models optimized for discriminative tasks such as MCQs tend to underperform in generative, semantic-heavy tasks, and vice versa. This inverse relationship suggests a fundamental architectural trade-off between structured knowledge retrieval and semantic text generation. General-purpose models such as LLaMA and Gemini demonstrate more balanced but less distinctive profiles, offering stable performance across tasks without excelling in any single dimension. While ChatDoctor demonstrates strong overall performance across multiple tasks, it performs relatively poorly on uncertainty reasoning as measured by PubMedQA. Conversely, Grok excels in areas where ChatDoctor is weaker, suggesting that these models have complementary strengths. Hence, these findings challenge the assumption of a universally optimal healthcare LLM and instead support a task-routing paradigm, where models are selected dynamically based on application needs. For rapid factual lookup, Grok-like systems are well suited; ChatDoctor-like models are more appropriate for clinical documentation and summarization; and LLaMA-style architectures show particular strength in reasoning under uncertainty.

The study also highlights the limitations of traditional evaluation metrics. While BLEU and ROUGE provide useful signals of linguistic overlap and content coverage, semantic metrics such as BERTScore better capture meaning preservation and clinical fidelity. Models with lower lexical overlap often achieved higher semantic alignment, underscoring that surface-level similarity alone is insufficient for evaluating medical text generation. Prompt consistency further played a critical role; sensitivity analyses showed that general-purpose models were more affected by minor wording changes, whereas domain-tuned models such as ChatDoctor remained more stable, which is an important consideration for real-world clinical deployment. From a safety and ethical perspective, ChatDoctor consistently adopted cautious, clinically appropriate language, explicitly expressing uncertainty when evidence was ambiguous, reinforcing the importance of conservative response behavior in safety-critical medical settings.

No single model excels across all healthcare tasks. Domain-tuned models offer superior semantic precision and safety alignment, while general-purpose models provide fluency, efficiency, and adaptability. Future healthcare AI systems should therefore adopt hybrid architectures that combine these complementary strengths. Moreover, although benchmark results are promising, real-world deployment will require further validation on noisy, heterogeneous clinical data alongside clinician-in-the-loop evaluation. LLMs should augment clinical judgment, supporting documentation, synthesis, and decision-making while maintaining human oversight and accountability.

\section*{Threats to Validity}

The experiments were conducted with limited computational resources, which constrained hyperparameter exploration and fine-tuning depth; larger-scale infrastructure could enable more detailed tuning, particularly for models such as LLaMA-3.1-8B, potentially enhancing output stability and clinical fidelity. Model outputs also showed some sensitivity to prompt formulation. Therefore, standardized prompts were used across all evaluations to ensure fair comparison, but future work could explore adaptive prompt strategies, few-shot conditioning, or ensemble approaches to further enhance robustness and reproducibility. The datasets used cover a broad spectrum of medical knowledge but primarily consist of curated or structured text, and may not fully capture the linguistic variability of real-world clinical documentation including abbreviations, fragmented phrasing, and institution-specific conventions. Likewise, evaluation metrics such as BLEU, ROUGE-L, and BERTScore provide meaningful indicators but cannot fully capture clinical correctness or safety. These considerations highlight opportunities for future work involving more diverse, multilingual, or de-identified clinical data and the inclusion of expert human evaluation to assess clinical reliability.

\section{Conclusion}
This study comparatively evaluated the performance of five LLMs across medical question answering and clinical summarization tasks, revealing that model effectiveness is inherently task-dependent. Our findings demonstrate that domain-specific models, such as ChatDoctor, exhibit superior contextual awareness and semantic reliability. Their cautious reasoning patterns and high alignment with clinical language make them uniquely suited for high-stakes medical applications where meaning preservation and patient safety are essential.

In contrast, general-purpose models like Grok and LLaMA excel in structured, fact-based question answering but struggle with the nuance and ambiguity typical of clinical judgment. This divergence suggests a fundamental trade-off: while general-purpose models offer broader adaptability and fluency, domain-tuned architectures provide the precision and risk-awareness necessary for clinical integration. Consequently, we conclude that no single architecture is universally optimal; rather, the selection of an LLM must be strictly aligned with the specific clinical demands of the task. Furthermore, this study advocates for a shift in evaluation paradigms. Traditional NLP metrics like BLEU and ROUGE proved insufficient for capturing clinical correctness or the gravity of medical errors. We propose that a multidimensional framework, integrating semantic measures with task-specific metrics, is essential for a meaningful assessment of medical AI. Such an approach not only captures linguistic quality and semantic fidelity but also accounts for contextual correctness and decision reliability. While LLMs hold considerable promise for enhancing healthcare delivery and decision support, their responsible integration depends on rigorous evaluation standards, domain-aware model selection, and sustained human involvement to ensure patient safety and ethical compliance.

% \section*{Declaration of conflicting interests}
% The authors declared no potential conflicts of interest with respect to the research, authorship, and/or publication of this article.

% \section*{Funding}
% The authors received no financial support for the research, authorship, and/or publication of this article.

% \section*{Ethics approval and consent to participate}
% Not Applicable

% \section*{Consent for publication}
% Not Applicable

\section*{Availability of data and materials}
The complete replication package, along with the medical datasets, is deposited into the GitHub Repository and is available at the following URL: https://github.com/subinsanthosh/Exploring-the-potential-of-LLMs-in-Healthcare.

% \section*{Competing interests}
% The authors declare that they have no competing interests and no potential conflicts of interest with respect to the research, authorship, and/or publication of this article.

% \section*{Funding}
% The authors received no financial support for the research, authorship, and/or publication of this article.

% \section*{Authors' contributions}
% All authors contributed to the study conception, design, analysis, and manuscript preparation, and approved the final version. S.S. worked on experimental evaluation and prepared figure 4-10. F.A. wrote the main manuscript and prepared figure 1, 2, 3, 11 and Table 1. H.A. worked on writing the manuscript especially the Discussion and Conclusion sections while also preparing Table 2, 3 and providing feedback on the draft. C.S. finalized the main concept of paper and worked on revising and refining the manuscript.

% \section*{Acknowledgements}
% The authors would like to thank all individuals and institutions who contributed to this study.
      
\bibliography{refy}

\end{document}